\documentclass[10pt,twocolumn,letterpaper]{article}

\usepackage{iccv}
\usepackage{times}
\usepackage{epsfig}
\usepackage{graphicx}
\usepackage{amsmath}
\usepackage{amssymb}
\usepackage{multirow}
\usepackage{mathabx}
\usepackage{arydshln}
\usepackage{booktabs}
\usepackage{utfsym}
\usepackage{caption}
\usepackage{algorithm}
\usepackage{algorithmic}
\usepackage{listings}
\usepackage{wrapfig}

\usepackage[breaklinks=true,bookmarks=false]{hyperref}

\iccvfinalcopy 

\def\iccvPaperID{****} 

\ificcvfinal\fi

\begin{document}

\title{Exploring Effective Factors for Improving Visual In-Context Learning}

\author{
Yanpeng Sun$^{1,2}$, Qiang Chen$^1$, Xiaofan Li$^1$, Jian Wang$^1$, Jingdong Wang$^1$, Zechao Li$^2$\thanks{Corresponding author.}\\
$^1$Baidu VIS\\
$^2$School of Computer Science and Engineering, Nanjing University of Science and Technology\\
}
\maketitle
\ificcvfinal\thispagestyle{empty}\fi

\begin{abstract}
The In-Context Learning (ICL) is to understand a new task via a few demonstrations (aka. prompt) and predict new inputs without tuning the models. While it has been widely studied in NLP, it is still a relatively new area of research in computer vision. To reveal the factors influencing the performance of visual in-context learning, this paper shows that prompt selection and prompt fusion are two major factors that have a direct impact on the inference performance of visual context learning. Prompt selection is the process of identifying the most appropriate prompt or example to help the model understand new tasks. This is important because providing the model with relevant prompts can help it learn more effectively and efficiently. Prompt fusion involves combining knowledge from different positions within the large-scale visual model. By doing this, the model can leverage the diverse knowledge stored in different parts of the model to improve its performance on new tasks. Based these findings, we propose a simple framework prompt-SelF for visual in-context learning. Specifically, we first use the pixel-level retrieval method to select a suitable prompt, and then use different prompt fusion methods to activate all the knowledge stored in the large-scale model, and finally ensemble the prediction results obtained from different prompt fusion methods to obtain the final prediction results. And we conduct extensive experiments on single-object segmentation and detection tasks to demonstrate the effectiveness of prompt-SelF. Remarkably, the prompt-SelF has outperformed OSLSM based meta-learning in 1-shot segmentation for the first time. This indicated the great potential of visual in-context learning. The source code and models will be available at \url{https://github.com/syp2ysy/prompt-SelF}.
\end{abstract}
\section{Introduction}

\begin{figure}
	\centering
	\includegraphics[width=\linewidth]{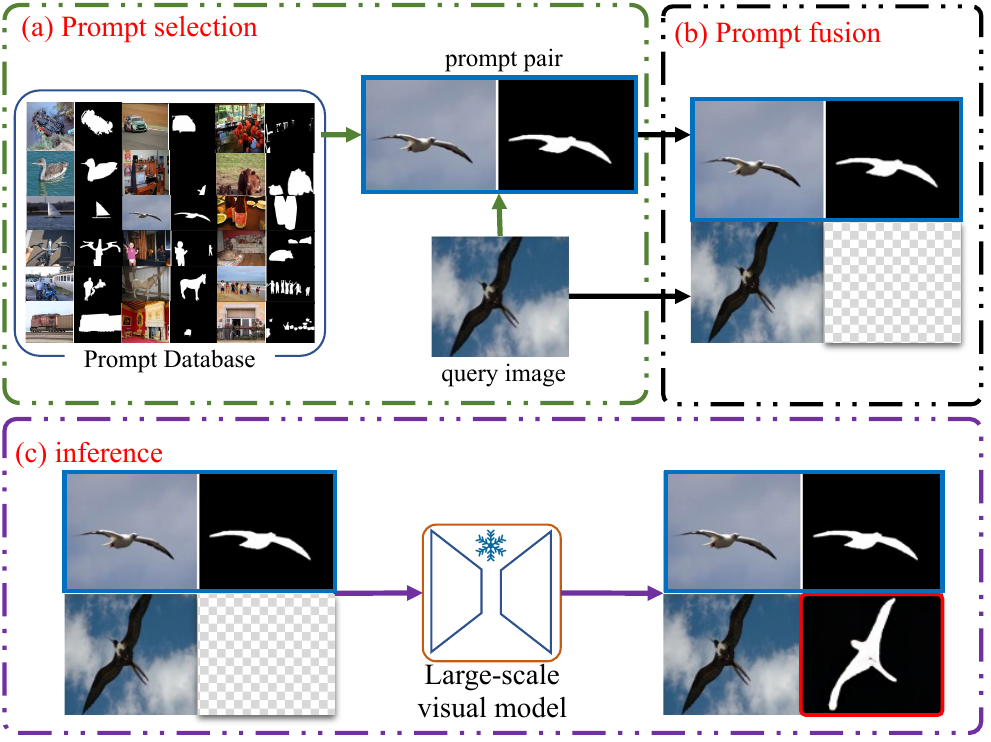}
	
	\caption{A framework for \textbf{visual in-context learning}. It involves the following steps. \textbf{(a)} Select the prompt pair corresponding to the task from the database based on the query image. \textbf{(b)} Fusion the query image and the selected prompt image to create a new image. \textbf{(c)} Feed new images into a pre-trained large-scale vision model to get predictions for query images. This paradigm enables the visual model to learn in-context by utilizing the knowledge captured from the prompt image to enhance the accuracy of the prediction for the demand image.}
	\label{fig:structure}
	\vspace{-0.3cm}
\end{figure}

 \begin{figure*}
	\centering
	\includegraphics[width=\linewidth]{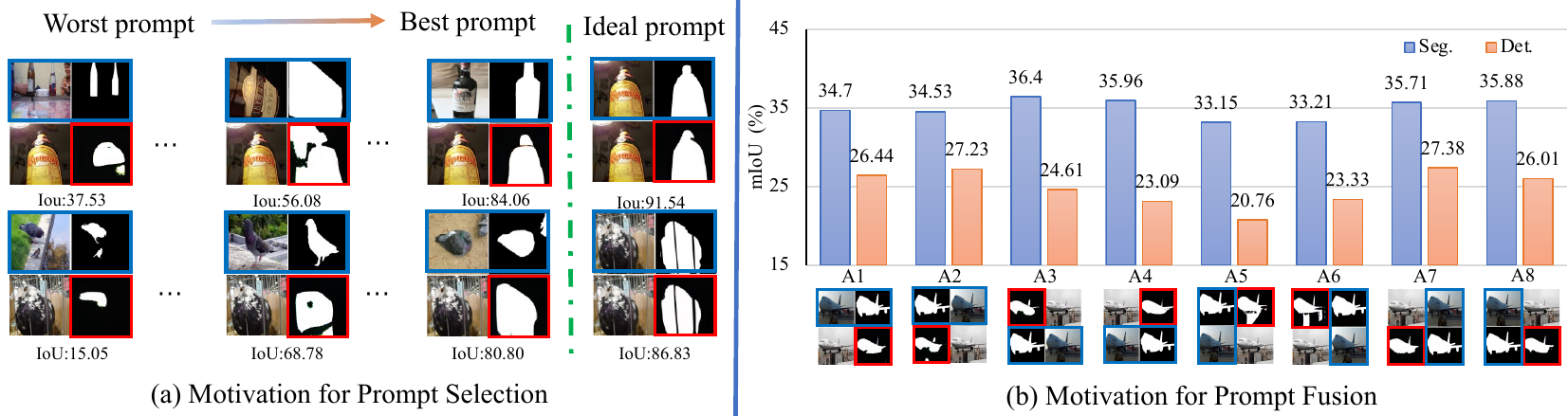}
	
	\caption{\textbf{(a)} The selection of different prompts can affect the results of the same query image, therefore, choosing a high-quality prompt can have a positive impact on visual in-context learning. \textbf{(b)} Different prompt fusion methods activate knowledge at different positions in the large-scale model, leading to varying degrees of impact on visual in-context learning. Note that the \textcolor{red}{red box} in the figure is the prediction result, and the \textcolor{blue}{blue box} is the prompt pair.}
	\label{fig:motivation}
	\vspace{-0.3cm}
\end{figure*}
Benefiting from the large models and large scale datasets in NLP, researchers realized that the large models~\cite{radford2018improving,radford2019language,brown2020language} have a crucial emergent ablity, which is In-context Learning. The purpose of in-context learning is to assist the model in comprehending new tasks and making predictions for new examples based on provided prompt. Typically, the prompt is a concise, structured input that provides context for the task, such as a task description or an example of an input-label pair. As a well-known field in NLP~\cite{gargcan,xieexplanation}, in-context learning has just started in the field of vision. Indeed, visual in-context learning is becoming increasingly important in computer vision, particularly with the rise of large-scale models. While these models can achieve impressive results on many tasks~\cite{he2022masked}, they often require massive amounts of data and computation to train, making them impractical for many real-world applications. As such, visual in-context learning is becoming increasingly important for developing more efficient and accurate computer vision systems that can operate in real-world settings. However, these research is relatively limited, so we are concentrating on visual in-context learning and carrying out preliminary studies.

Different from NLP, where the input is typically text, the input in visual tasks is often a single image. Therefore, the challenge of visual in-context learning lies in how to make the model understand the prompt and make predictions on the new image (aka. query image). Toward this end, recent studies~\cite{wang2022images,barvisual} propose unified visual in-context learning frameworks, illustrated in Figure~\ref{fig:structure}. It involve selecting a prompt from a database based on query image, fusing it with the query image to create a combine images, and feeding this image into a large-scale model to obtain a prediction. Recent studies~\cite{rubin2022learning,min2022rethinking,min2022metaicl} have demonstrated that the quality of prompts is the primary determinant of performance in language in-context learning. In the context of visual in-context learning, we ask: \textit{what are the critical factors that influence downstream task performance?}

Under the premise of large-scale model determination, cue selection and fusion are the main determinants of visual context learning performance. We first analyze the prompt selection. Prior work in NLP~\cite{liu2021makes} has shown that a high-quality prompt can help the model better understand the task and yield improved results. We found that when different prompts were used for the same query image (as shown in Figure~\ref{fig:motivation} (a)), the prediction results were highly sensitive to the prompt selection. A high-quality prompt can lead to a significantly increase in Intersection over Union (IoU) for the prediction result. Thus, one of the challenges in visual in-context learning is how to efficiently select high-quality prompt. In Figure~\ref{fig:motivation} (a), we obtained the best results by using the query image itself as the prompt, indicating that high-quality prompts are spatially similar to the query image. For prompt selection, vpr~\cite{zhang2023makes} propose an image-level retrieval framework. However, this approach does not account for certain characteristics of high-quality prompts.

Prompt fusion aims to merge the prompt and query image into a new image, which becomes the input for the large-scale model. The new image includes four sub-images: the prompt image, prompt ground truth, query image, and prediction result. To ensure that the prompt pair is combined correctly (adjacent to each other), prompt fusion has a total of eight different arrangements (as shown in Figure~\ref{fig:motivation} (b)). In this paper, we conduct a comprehensive investigation into the impact of different sub-image arrangements on in-context learning, and identify a critical issue: downstream performance is highly sensitive to variations in sub-image arrangements. Because most large-scale vision models are based on masked image modeling (MIM)~\cite{he2022masked,barvisual,chen2022context,baobeit}, different prompt fusion methods activate knowledge at different locations in the large-scale model, affecting the downstream task performance. Thus, determining which prompt fusion method is the most effective for new downstream tasks is one of the major challenges in visual context learning. 

Toward this end, we propose a simple framework prompt-SelF for visual in-context learning. In prompt-SelF, we first use the pixel-level retrieval method to select a suitable prompt image that is similar to the input image in terms of visual content and context. This is important because a good prompt can provide relevant information and guidance to the model for understanding the new task. Once a suitable prompt image is selected, we then use different prompt fusion methods to activate all the knowledge stored in the large-scale model. This involves combining the prompt image with the input image in various ways to create a composite image that can better capture the relevant information needed for the task. By doing this, we can activate diverse knowledge in the large-scale model and improve the accuracy and efficiency of in-context learning. Finally, we ensemble the prediction results obtained from different prompt fusion methods to obtain the final prediction results. This is done by using a majority voting strategy~\cite{lewkowyczsolving}. The ensemble strategy helps to further improve the performance of visual in-context learning by combining the strengths of different prompt fusion methods. We have conducted extensive experiments on single-object segmentation and detection tasks using visual in-context learning. Notably, our proposed approach, prompt-SelF, achieved state-of-the-art results and outperformed meta-learning-based methods for the first time on single-object segmentation tasks.

\section{Related work}
\subsection{In-context learning}
With the emergence of large models such as GPT-3~\cite{brown2020language} and Instruction GPT~\cite{ouyangtraining}, prompting these models more efficiently has become a concern for both academia and industry. As a result, the in-context learning method has gained popularity in the field of NLP. In-context learning aims to improve the performance of these models by providing additional context during inference. In recent years, in-context learning has been applied to a variety of NLP tasks~\cite{garcia2022using,min2022metaicl}, including language modeling~\cite{kenton2019bert,brown2020language}, text classification~\cite{rubin2022learning,min2022rethinking}, question answering~\cite{press2022measuring,liu2021makes,khot2020qasc}, and natural language generation~\cite{kim2022self}. 

However, the concept of in-context learning is still relatively new in computer vision~\cite{alayracflamingo}. The first step in implementing in-context learning is to unify different tasks into the same space. However, while some recent work~\cite{kolesnikovuvim,chenpix2seq,lu2022unified} has attempted to unify different tasks, they have been limited in their ability to learn new tasks. To solve this problem, some recent works~\cite{wang2022images,barvisual,zhang2023makes} activate the model's ability to learn new tasks by combining prompt and query image fusion together. But it does not explore the factors that affect the performance of visual in-context learning. Therefore, we have conducted the first exploration of the main factors that affect the performance of visual in-context learning inference.

\subsection{Optimization method of ICL in NLP}
In the inference stage, optimizing ICL aims to improve the performance of the model on new tasks without altering the large-scale model~\cite{ouyangtraining}. The prompt, as an input to activate the ability of large models, has a significant impact on ICL. In natural language processing, prompts are often designed based on organization and format~\cite{dong2022survey}. The organization method refers to how to select a suitable prompt. Existing work~\cite{kim2022self,zhang2022active,lu2022fantastically} can select similar results through text representation and mutual information, and prompts can even be directly generated by the language model itself. The format of the prompt can be divided into instruction~\cite{ouyangtraining,wang2022benchmarking} and reasoning steps~\cite{weichain,wang2022self}. The instruction is mainly designed manually, and the reasoning step is designed to inspire the reasoning ability of the model by introducing reasoning steps. To advance the field of visual in-context learning, this paper presents the first exploration of optimization methods for visual in-context learning.
\section{Methods}
In this section, we start with the preliminaries on the visual in-context learning setting. Then, the analysis of two factors involved in visual in-context learning is presented, followed by a detailed introduction of the prompt-SelF process. Specifically, Prompt selection and Prompt fusion are discussed in sections~\ref{sec3.2} and section~\ref{sec3.3}, respectively.

\subsection{Visual In-context learning}
In natural language processing, remarkable progress has been made in the development of in-context learning. Large-scale pre-trained language models, such as Instruction GPT~\cite{ouyangtraining}, have introduced text prompts into the input, allowing the model to predict and learn different tasks without the need to update parameters. In computer vision, the concept of in-context learning is still relatively new. Previous studies~\cite{wang2022images,barvisual} have focused on combining images and labels into a new image and using MIM for pre-training, resulting in models with in-context learning capabilities. However, their performance on new tasks was not satisfactory. The primary goal of this paper is to investigate the key factors that impact visual context learning and improve the current frameworks. 

In new task, given a prompt dataset $C = \left\{ (x_1,y_1), (x_2,y_2), .... , (x_n,y_n) \right\}$, in which there are $n$ pairs of image-label pairs, $(x_k,y_k)$ represents the $k-th$ image-label pair. If we have an image $x_q$ and a model $f_{icl} $ with in-context learning capability, then the process of in-context learning can be expressed as follows:\\
\begin{equation}
  y_q = f_{icl}^{*}(x_q | (x_i,y_i)) 
\end{equation}
where, $f_{icl}^{*}$ represents the visual model $f_{icl} $ with frozen parameters. $(x_i,y_i)$ refers to the prompt pair selected from the prompt dataset $C$ based on query image $x_q$.

Therefore, changing prompt has a direct impact on the performance of visual context learning. Additionally, the arrangement of prompt and query images is not fixed, and changing the arrangement can activate different knowledge from various positions in the large-scale visual model. As a result, we can conclude that prompt selection and prompt fusion are the main factors that affect visual in-context learning. Based on this analysis, we have designed a simple framework called Prompt-SelF.

\subsection{Prompt Selection}\label{sec3.2}
The purpose of prompt selection is to automatically select a high-quality prompt for the query image. We ask, \textit{what are the characteristics of a high-quality prompt?} The impact of different prompts on the prediction results of the query image is illustrated in Figure~\ref{fig:motivation} (a). Our findings suggest that prompts with high spatial similarity to the query image can have a positive impact on the prediction results. Based on this finding, we propose the Prompt-SelF framework that leverages pixel-level retrieval to identify the suitable prompt for the query image. 

The detailed process of prompt selection based pixel-level retrieval is shown in Algorithm~\ref{alg:ps}. First, we use the off-the-shelf feature extractor to obtain the spatial representation $F_q$ of the query image and the spatial representation $F_s$ of all images in the dataset $C$. Then, we normalize the image features using L2 normalization and calculate the pixel-level similarity between  the query image and all images in the dataset following form. \\
\begin{equation}
    x_P = arg \max_{x_i\in C} (norm(F_q) \cdot norm(F_s)^{\top }  )
\end{equation}
Finally, we select the image-label pair with the highest similarity score as the prompt $(x_P, y_P)$. Once we have retrieved the appropriate prompt for the query image, the next step is to fuse the prompt and the query image, and use the large-scale visual model to make predictions.
\begin{figure}
\centering
\begin{minipage}{1.0\linewidth}
\begin{algorithm}[H]\small
\caption{Pseudo code of Prompt Selection}
\label{alg:ps}
\definecolor{codeblue}{rgb}{0.25,0.5,0.5}
\definecolor{codekw}{rgb}{0.85, 0.18, 0.50}
\lstset{
  backgroundcolor=\color{white},
  basicstyle=\fontsize{7.5pt}{7.5pt}\ttfamily\selectfont,
  columns=fullflexible,
  breaklines=true,
  captionpos=b,
  commentstyle=\fontsize{7.5pt}{7.5pt}\color{codeblue},
  keywordstyle=\fontsize{7.5pt}{7.5pt}\color{codekw},
}
\begin{lstlisting}[language=python]
# Input: all images in the dataset 'X', query image 'x_q', the off-the-shelf feature extractor 'img_encoder'
# Output: Output prompt index 'idx'

F_s = img_encoder.forward(X) # (N,C,H,W)
F_q = img_encoder.forward(x_q) # (1,C,H,W)

# normalization & reshape
F_s = (F_s / norm(F_s, dim=1)).reshape(N,-1)
F_q = F_q / norm(F_q, dim=1).reshape(1,-1)

# calculate similarity & get index
similarity = dot(F_q, F_s)  # (1,N)
idx = argsort(similarity,dim=1)[0]
\end{lstlisting}
\end{algorithm}
\end{minipage}
\vspace{-0.5cm}
\end{figure}

\subsection{Prompt fusion}\label{sec3.3}
The purpose of prompt fusion is to combine the prompt and query image into a new input image for the large-scale model, to obtain the predicted result for the query image. There are eight possible ways to arrange the sub-images while ensuring that the prompt pair is adjacent to each other. Different arrangements have varying effects on visual in-context learning, as illustrated in Figure~\ref{fig:motivation} (b). The reason for this phenomenon is that different arrangements of sub-images activate different parts of the knowledge stored in the large-scale model. We ask, \textit{How can we leverage the knowledge stored in the large-scale model to its fullest potential?} Toward this end, Prompt-SelF proposes a simple method to maximizing the potential of large-scale vision models by employing different permutations of sub-images to produce multiple predictions, which are then merged to yield more precise outcomes.

Algorithm~\ref{alg:pf} provides a detailed overview of the prompt fusion process in Prompt-SelF. It aims to fully leverage the diverse knowledge in large-scale visual models. To achieve this, we fuse the query image and prompt image using different arrangements to create eight new fused images, as illustrated in Figure~\ref{fig:motivation} (b). These eight new images are then input into the frozen large-scale visual model to obtain eight prediction results based on the knowledge of different positions in the model. Finally, we ensemble these eight results to obtain the final prediction. In the final stage of ensemble, we use the simple voting strategy, where the default threshold in the paper is set to $4/8$. This means that if more than four out of the eight predictions are in agreement, the final prediction is made based on their consensus.

In summary, we analyzed the main factors that affect visual in-context learning and presented the implementation details of the proposed prompt-SelF framework. By selecting higher-quality prompts and leveraging the diverse knowledge captured by large-scale visual models, prompt-SelF can effectively improve the accuracy and efficiency of contextual learning. Subsequently, we conducted experiments to evaluate the effects of prompt selection and prompt fusion on visual in-context learning, and to validate the efficacy of Prompt-SelF.

\begin{figure}
\centering
\begin{minipage}{1.0\linewidth}
\begin{algorithm}[H]\small
\caption{Pseudo code of Prompt Fusion}
\label{alg:pf}
\definecolor{codeblue}{rgb}{0.25,0.5,0.5}
\definecolor{codekw}{rgb}{0.85, 0.18, 0.50}
\lstset{
  backgroundcolor=\color{white},
  basicstyle=\fontsize{7.5pt}{7.5pt}\ttfamily\selectfont,
  columns=fullflexible,
  breaklines=true,
  captionpos=b,
  commentstyle=\fontsize{7.5pt}{7.5pt}\color{codeblue},
  keywordstyle=\fontsize{7.5pt}{7.5pt}\color{codekw},
}
\begin{lstlisting}[language=python]
# Input: prompt pair '(x_p,y_p)', query image 'x_q', the visual generative model 'gen_model'
# Output: Output prediction result 'pred_q'

#Get the fusion image of 8 arrangements
arr_list = get_fusion_image(x_p,y_p,x_q)

#Predict results on 8 fusion images
pred_list = []
for fusion_image in arr_list:
    sub_pred = gen_model.forward(fusion_image)
    pred_list.append(sub_pred)
    
#get the final prediction result
pred_q = ensembel(pred_list)
\end{lstlisting}
\end{algorithm}
\end{minipage}
\vspace{-0.5cm}
\end{figure}

\vspace{-0.5em}
\section{Experiments}
In this section, we perform comprehensive experiments on single-object segmentation and detection tasks to validate several crucial factors that affect visual context learning and demonstrated the effectiveness of prompt-SelF. In section~\ref{section4.1}, we introduce the experimental setup and implementation details. In section~\ref{section4.2}, we compare our proposed prompt-SelF method with state-of-the-art methods to verify its superiority and versatility. Finally, in section~\ref{section4.3}, we analyze the characteristics of prompt-SelF to further demonstrate its effectiveness.

\subsection{Setting}\label{section4.1}
\noindent \textbf{Downstream tasks and Datsets.} To ensure fair comparison, we adopt the experimental settings of previous works~\cite{barvisual,zhang2023makes} for single-object segmentation, single object detection. Next, we will provide a detailed introduction to these tasks and the datasets used.

\begin{itemize}
\item \textbf{single object segmentation}: The goal is to extract the specific object from the query image based on the prompt. This experimental setup is similar to few-shot segmentation, where new object classes in the query image are recognized based on few of labeled samples. Thus, following previous research~\cite{barvisual,sunsingular,wang2019panet,li2021adaptive}, we evaluate our method on the pasacl-5$^i$~\cite{shaban2017one} dataset.
\item \textbf{single object detection}: Different from traditional detection tasks, the goal here is to obtain fine-grained segmentation results based on coarse-grained bounding box annotations in prompt. Following the previous work~\cite{barvisual}, we conduct experiments on images and bounding boxes from the PASCAL VOC 2012~\cite{everingham2009pascal} dataset, specifically selecting images that contain only a single object and filtering out trivial images where the object covers more than 50\% of the image.
\end{itemize}

\begin{table}[t]\footnotesize
\centering
\setlength\tabcolsep{2.5pt}
\renewcommand\arraystretch{1.3}
\caption{Main results on different computer vision tasks. The best results based in-context learning are show in \textbf{bold}.}
\label{tab:main_results}
\begin{tabular}{cccccccc}
\hline
\multicolumn{1}{c|}{}            & \multicolumn{5}{c|}{Seg.(mIoU)$\uparrow$}                             & \multicolumn{1}{c}{\multirow{2}{*}{\begin{tabular}[c]{@{}c@{}}Det.\\ (mIoU)$\uparrow$\end{tabular}}}  \\
\multicolumn{1}{c|}{}            & Fold-0 & Fold-1 & Fold-2 & Fold-3 & \multicolumn{1}{c|}{Mean}  &                                     \\ \hline
\multicolumn{7}{c}{\textit{Meta-learning}}                                                                                                                                           \\ \hline
\multicolumn{1}{c|}{OSLSM \cite{shaban2017one}}       & 33.60   & 55.30   & 40.90   & 33.50   & \multicolumn{1}{c|}{40.80} & \multicolumn{1}{c}{-}                          \\
\multicolumn{1}{c|}{co-FCN \cite{rakelly2018conditional}}      & 36.70   & 50.60   & 44.90   & 32.40   & \multicolumn{1}{c|}{41.10} & \multicolumn{1}{c}{-} \\ \hline
\multicolumn{7}{c}{\textit{In-context learning}}                                                                                                                                    \\ \hline
\multicolumn{1}{c|}{VP-Random $^{\text{\textdagger}}$ \cite{barvisual}}      & 27.49   & 30.21   & 25.88   & 24.06   & \multicolumn{1}{c|}{26.91} & \multicolumn{1}{c}{25.14}                      \\
\multicolumn{1}{c|}{VPR-UsupPR $^{\text{\textdagger}}$ \cite{zhang2023makes}} & 34.70   & 35.92   & 32.39   & 31.12   & \multicolumn{1}{c|}{33.53} &  \multicolumn{1}{c}{26.44}                    \\

\multicolumn{1}{c|}{VPR-SupPR $^{\text{\textdagger}}$ \cite{zhang2023makes}} & 37.08   & 38.43   & 34.40   & 32.32   & \multicolumn{1}{c|}{35.56} &  \multicolumn{1}{c}{28.22}                     \\ \hline
\multicolumn{1}{c|}{prompt-SelF}        & \textbf{42.48}   & \textbf{43.34}   & \textbf{39.76}   & \textbf{38.50}   & \multicolumn{1}{c|}{\textbf{41.02}} & \multicolumn{1}{c}{\textbf{29.83}}                  \\ \hline
\end{tabular}
\vspace{-0.3cm}
\end{table}

\noindent \textbf{Evaluation method.} We conducted a comprehensive evaluation of the effectiveness of prompt-SelF by comparing it with several existing methods for learning visual context, including VP~\cite{barvisual} and VPR~\cite{zhang2023makes}. 

Additionally, we compared prompt-SelF with few-shot segmentation methods  based  meta-learning~\cite{shaban2017one, rakelly2018conditional}. In the ablation experiment, we compared three prompt selection methods, namely random selection, image-level retrieval, and pixel-level retrieval. Moreover, we evaluated the impact of different prompt arrangements, denoted as Arrangement1 to Arrangement8, which correspond to A1-A8 in Figure~\ref{fig:motivation} (b). We conduct this comparison to validate the significance of prompt selection and prompt fusion in visual in-context learning, while also ensuring a comprehensive and impartial assessment of the effectiveness of prompt-SelF.

\noindent \textbf{Implementation details.} To perform in-context learning, not all large-scale models in computer vision are capable enough. Thus, we employ a pre-trained model named MAE-VQGAN~\cite{barvisual} specifically designed for this purpose. In the prompt selection phase, we extract features using CLIP's visual encoder~\cite{radford2021learning}, which yields two types of features: image-level features and pixel-level features. The former is obtained by using the classification header, while the latter is obtained without it. Since the pre-training model's input size is $224 \times224$, we maintain the same size for image inputs in our experiments. We divide the image into sub-images of size $111 \times111$, with a spacing of $1$ pixel between each sub-image.

\begin{table}[]\footnotesize
\centering
\setlength\tabcolsep{2.5pt}
\renewcommand\arraystretch{1.3}
\caption{Ablation study of different prompt selection retrieval methods. \protect\usym{2717} and \protect\usym{2713} represent the prompt fusion method of \textbf{A1} and \textbf{prompt-SelF} respectively. The best results based in-context learning are show in \textbf{bold}.}
\label{tab:diff_retrieval}
\begin{tabular}{c|c|cccc|c}
\hline
prompt selection                       & prompt fusion & Fold-0 & Fold-1 & Fold-2 & Fold-3 & Means \\ \hline
\multirow{2}{*}{random}      &   \usym{2717}       & 27.49 & 30.21 & 25.88  & 24.06 & 26.91 \\
                             &    \usym{2713}      & 33.91 & 36.12 & 31.97  & 29.37 & 32.84\\ \hline
\multirow{2}{*}{image-level} &    \usym{2717}      & 34.70  & 35.92  & 32.92  & 31.12  & 33.53 \\
                             &   \usym{2713}       & 41.07  & 41.32  & 38.14  & 36.44  & 39.24 \\ \hline
\multirow{2}{*}{pixel-level} &    \usym{2717}      & 36.42  & 38.47  & 34.56  & 34.12  & \textbf{35.89} \\
                             &    \usym{2713}      & 42.48  & 43.34  & 39.76  & 38.50  & \textbf{41.02} \\ \hline
\end{tabular}
\vspace{-0.4cm}
\end{table}

\begin{table*}[]
 \caption{Ablation study of different sub-images arrangements. \textcolor{cyan}{Image-level} and \textcolor{magenta}{Pixel-level} represent prompt selection methods based on image-level retrieval and pixel-level retrieval, respectively. The best results based in-context learning are show in \textbf{bold}.}
 \label{tab:diff_position}
 \centering
 \renewcommand\arraystretch{1.3}
 \setlength{\tabcolsep}{3.4mm}{
 \resizebox{1.0\linewidth}{!}{
 \begin{tabular}{c|ccccc|ccccc}
\hline
           & \multicolumn{5}{c|}{\textcolor{cyan}{Image-level}}                               & \multicolumn{5}{c}{\textcolor{magenta}{Pixel-level}}                                  \\ \cline{2-11} 
           & Fold-0 & Fold-1 & Fold-2 & \multicolumn{1}{c|}{Fold-3} & Mean  & Fold-0 & Fold-1 & Fold-2 & \multicolumn{1}{c|}{Fold-3} & Mean  \\ \hline
Arrangement1 & 34.70  & 35.92  & 32.92  & \multicolumn{1}{c|}{31.12}  & 33.53 & 36.42  & 38.47  & 34.56  & \multicolumn{1}{c|}{34.12}  & 35.89 \\
Arrangement2 & 34.53  & 35.82  & 32.20  & \multicolumn{1}{c|}{31.29}  & 33.46 & 36.46  & 38.49  & 34.30  & \multicolumn{1}{c|}{33.81}  & 35.77 \\
Arrangement3 & 36.40  & 38.06  & 34.19  & \multicolumn{1}{c|}{34.07}  & 35.68 & 37.93  & 40.48  & 35.90  & \multicolumn{1}{c|}{36.07}  & 37.60 \\
Arrangement4 & 35.96  & 37.53  & 33.63  & \multicolumn{1}{c|}{32.95}  & 35.02 & 36.70  & 39.62  & 35.62  & \multicolumn{1}{c|}{34.87}  & 36.70 \\ \hline
Arrangement5 & 33.15  & 38.25  & 31.83  & \multicolumn{1}{c|}{32.39}  & 33.91 & 33.55  & 39.21  & 33.86  & \multicolumn{1}{c|}{34.80}  & 35.36 \\
Arrangement6 & 33.21  & 36.01  & 31.31  & \multicolumn{1}{c|}{32.18}  & 33.18 & 34.57  & 38.54  & 33.12  & \multicolumn{1}{c|}{34.37}  & 35.15 \\
Arrangement7 & 35.71  & 36.81  & 35.49  & \multicolumn{1}{c|}{30.77}  & 34.70 & 37.89  & 39.24  & 37.16  & \multicolumn{1}{c|}{34.16}  & 37.11 \\
\multicolumn{1}{l|}{Arrangement8} &
  35.88 &
  38.29 &
  36.43 &
  \multicolumn{1}{c|}{31.98} &
  35.65 &
  \multicolumn{1}{l}{37.68} &
  \multicolumn{1}{l}{39.80} &
  \multicolumn{1}{l}{37.65} &
  \multicolumn{1}{l|}{34.86} &
  \multicolumn{1}{l}{37.50} \\ \hline
\multicolumn{1}{c|}{ensemble} &
  \textbf{41.07} &
  \textbf{41.32} &
  \textbf{38.14} &
  \multicolumn{1}{c|}{\textbf{36.44}} &
  \textbf{39.24} &
  \multicolumn{1}{l}{\textbf{42.48}} &
  \multicolumn{1}{l}{\textbf{43.34}} &
  \multicolumn{1}{l}{\textbf{39.76}} &
  \multicolumn{1}{l|}{\textbf{38.50}} &
  \multicolumn{1}{l}{\textbf{41.02}} \\ \hline
\end{tabular}
}
}
\vspace{-0.4em}
\end{table*}

 \begin{figure*}
	\centering
	\includegraphics[width=\linewidth]{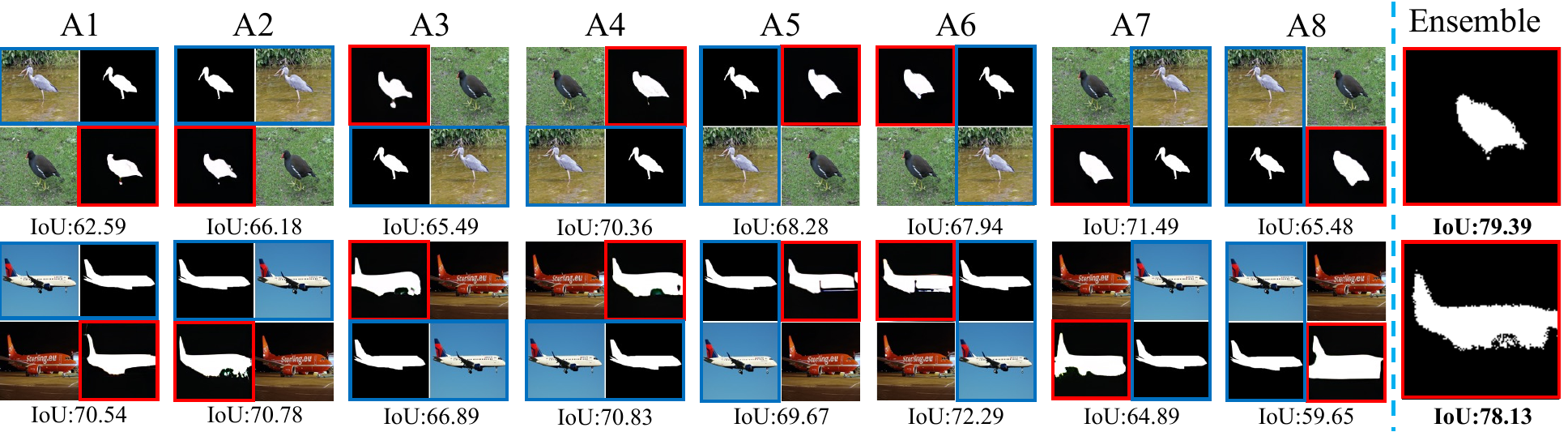}
	
	\caption{Visualization results of different prompt fusion methods. A1-A8 respectively correspond to Arrangement1-Arrangement8 in Table~\ref{tab:diff_position}, where the \textcolor{blue}{blue box} is the prompt pair, and the \textcolor{red}{red box} is the prediction result. The best results are show in \textbf{bold}.}
	\label{fig:vis_arra}
	\vspace{-0.4cm}
\end{figure*}

\vspace{-0.3em}
\subsection{Comparison with State-of-the-Art}\label{section4.2}
To evaluate the superiority of prompt-SelF, we compared it not only with the existing in-context learning method, but also with a few-shot segmentation method~\cite{shaban2017one,rakelly2018conditional} based on meta-learning. The experimental results are presented in Table~\ref{tab:main_results}, indicating that our method significantly enhances the performance of visual in-context learning across different tasks. It's important to highlight that prompt-SelF has demonstrated superior performance compared to an unsupervised method VPR-UsupPR~\cite{zhang2023makes} in different downstream tasks, particularly achieving a substantial $7.5$ mIoU improvement in segmentation task. Additionally, our method also outperforms VPR-SupPR~\cite{zhang2023makes} that necessitate training. The remarkable performance of prompt-SelF in the segmentation task, surpassing that of meta-learning based methods for the first time, is particularly surprising. These results showcase both the superiority of prompt-SelF and the immense potential of visual context learning. By selecting high-quality prompts and leveraging the diverse knowledge captured by large-scale vision models, we can significantly enhance the accuracy and efficiency of context learning, leading to further advancements in the field of visual recognition and analysis.

\vspace{-0.3em}
\subsection{Ablation Study}\label{section4.3}
In this section, we conduct an analysis of the factors that affect visual in-context learning and verify the effectiveness of prompt-SelF. All experiments below are performed on single-object segmentation tasks.\\

\vspace{-0.5em}
\noindent \textbf{Prompt selection.} It aim to select the best prompt for a query image, which can be done using various methods, such as random selection~\cite{barvisual} or retrieval based on image-level features~\cite{zhang2023makes}. However, based on the analysis presented in Figure~\ref{fig:motivation} (a), we identified the characteristic of high-quality prompts and proposed prompt-SelF, a retrieval method based on pixel-level features. To validate the effectiveness of this method, we compared it with the other two prompt selection methods. The random retrieval setting is identical to that of VP~\cite{barvisual}. Based on the results presented in Table ~\ref{tab:diff_retrieval}, we observe a significant improvement in the performance of single object segmentation task when using hints obtained by pixel-level retrieval. This suggests that the prompts obtained through the retrieval method based on pixel-level features are better suited to the characteristics of high-quality prompts. Additionally, we also observed that changing the prompt fusion method did not compensate for the negative impact of prompt quality. These results demonstrate that prompt selection significantly impacts visual in-context learning, and prompt fusion is also an independent factor that affects visual in-context learning.\\
\vspace{-0.1cm}

\begin{table*}[htp]
 \centering
 \renewcommand\arraystretch{1.2}
\caption{Ablation study of domain shift on single object segmentation. \protect\usym{2717} and \protect\usym{2713} represent the prompt fusion method of A1 and prompt-SelF respectively. \textit{Pascal $\rightarrow$ Pascal} and \textit{COCO $\rightarrow$ Pascal} mean from Pascal to Pascal and from COCO to Pascal. The best results based in-context learning are show in \textbf{bold}.}
\label{tab:domain}
 \setlength{\tabcolsep}{3.6mm}{
 \resizebox{1.0\linewidth}{!}{
\begin{tabular}{c|c|c|cccc|c}
\hline
Prompt selection                       &  Domain shift   & Prompt fusion          & Fold-0 & Fold-1 & Fold-2 & Fold-3 & Means \\ \hline
\multirow{4}{*}{\textcolor{cyan}{Image-level}} & \textit{Pascal $\rightarrow$ Pascal} & \multirow{2}{*}{\usym{2713}} & 34.70  & 35.92  & 32.92  & 31.12  & 33.53 \\
                             & \textit{COCO$\rightarrow$ Pascal} &                   & 33.05  & 34.91  & 28.92  & 30.62  & 31.88\textcolor{red}{$_{(-1.65)}$} \\ \cline{2-8} 
                             & \textit{Pascal $\rightarrow$ Pascal} & \multirow{2}{*}{\usym{2717}} & 41.07  & 41.32  & 38.14  & 36.44  & 39.24 \\
                             & \textit{COCO$\rightarrow$ Pascal} &                   & 39.31  & 40.17  & 34.80  & 35.75  & 37.35\textcolor{red}{$_{(-1.89)}$} \\ \hline
\multirow{4}{*}{\textcolor{magenta}{Pixel-level}} & \textit{Pascal $\rightarrow$ Pascal} & \multirow{2}{*}{\usym{2713}} & 36.42  & 38.47  & 34.56  & 34.12  & 35.89 \\
                             & \textit{COCO$\rightarrow$ Pascal} &                   & 33.94  & 37.45  & 32.45  & 33.21  & 34.26\textcolor{red}{$_{(-1.62)}$} \\ \cline{2-8} 
                             & \textit{Pascal $\rightarrow$ Pascal} & \multirow{2}{*}{\usym{2717}} & 42.48  & 43.34  & 39.76  & 38.50  & \textbf{41.02} \\
                             & \textit{COCO$\rightarrow$ Pascal} &                   & 40.13  & 42.14  & 37.84  & 38.52  & \textbf{39.80}\textcolor{red}{$_{(-1.22)}$}  \\ \hline
\end{tabular}
}
}
\end{table*}

\vspace{-0.3em}
\noindent \textbf{Prompt fusion.} The purpose of prompt fusion is to combine the prompt and query image into a new image, which is then fed into a large-scale visual model for prediction. Previous work followed the combination method shown as A1 in Figure~\ref{fig:motivation} (b). However, there are a total of eight possible sub-images arrangements while ensuring that the prompt pair is adjacent. Next, we compared these eight arrangements while using the same prompt selection methods to evaluate the impact of different arrangements on visual context learning. The results are presented in Table ~\ref{tab:diff_position}, indicating that variations in the arrangement of sub-images have varying effects on visual in-context learning. We hypothesize that the reason for this is the variation in the knowledge representation in different positions of the large-scale model. To substantiate this, we provide the visualization results of different arrangements, as depicted in Figure~\ref{fig:vis_arra}. The visualization outcomes demonstrate that it is arduous to establish the best arrangement for each query image. Moreover, a solitary sub-images arrangement in visual in-context learning fails to leverage the complete potential of knowledge present in large-scale visual models. Hence, Prompt-SelF proposes a straightforward ensemble strategy. We will discuss the ensemble strategy in prompt fusion in the following section.\\

\begin{figure}
	\centering
	\includegraphics[width=\linewidth]{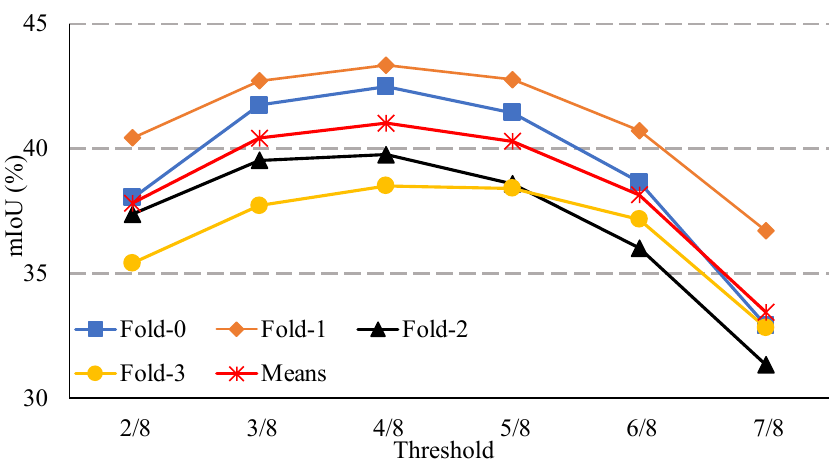}
	
	\caption{Ablation study of thresholds in ensemble strategy. The denominator and numerator of the threshold indicate the complete count of arrangements and the minimum count needed to establish the final outcome, correspondingly.}
	\label{fig:thresholds}
	\vspace{-0.7cm}
\end{figure}

\vspace{-0.3em}
\noindent \textbf{Ensemble strategy.} To fully exploit the knowledge in large-scale visual models, we propose a simple ensemble strategy. This strategy integrates the prediction results of eight different arrangements and uses a simple voting strategy to determine the final result. Table~\ref{tab:diff_position} presents the experimental results, which demonstrate that the ensemble strategy can significantly enhance the performance of visual in-context learning. This suggests that the ensemble strategy is effective in leveraging the diverse knowledge captured by the large-scale visual model and leads to more accurate predictions. Furthermore, we observed that the prompt selection strategy does not affect the success of the ensemble strategy. The visualizations presented in Figure~\ref{fig:vis_arra} illustrate that the ensemble's prediction outcomes can achieve more comprehensive object segmentation. 

We have set a decision threshold in our ensemble strategy to determine the final result. For instance, when the threshold is 2/8, the pixel will be classified as belonging to a certain class if at least two of the eight predictors predict the same class for the pixel. Next, we conducted ablation study on different thresholds to determine the optimal threshold value for Prompt-SelF. The experimental results are shown in Figure~\ref{fig:thresholds}, which provides insights into the effect of varying threshold values on the performance of our ensemble strategy. Our experiments reveal that the optimal threshold value for the majority voting ensemble strategy is $4/8$, as it provides the best performance on segmentation tasks. Therefore, in this paper, we set the threshold for ensemble to $4/8$ by default.\\

 \begin{figure*}
	\centering
	\includegraphics[width=\linewidth]{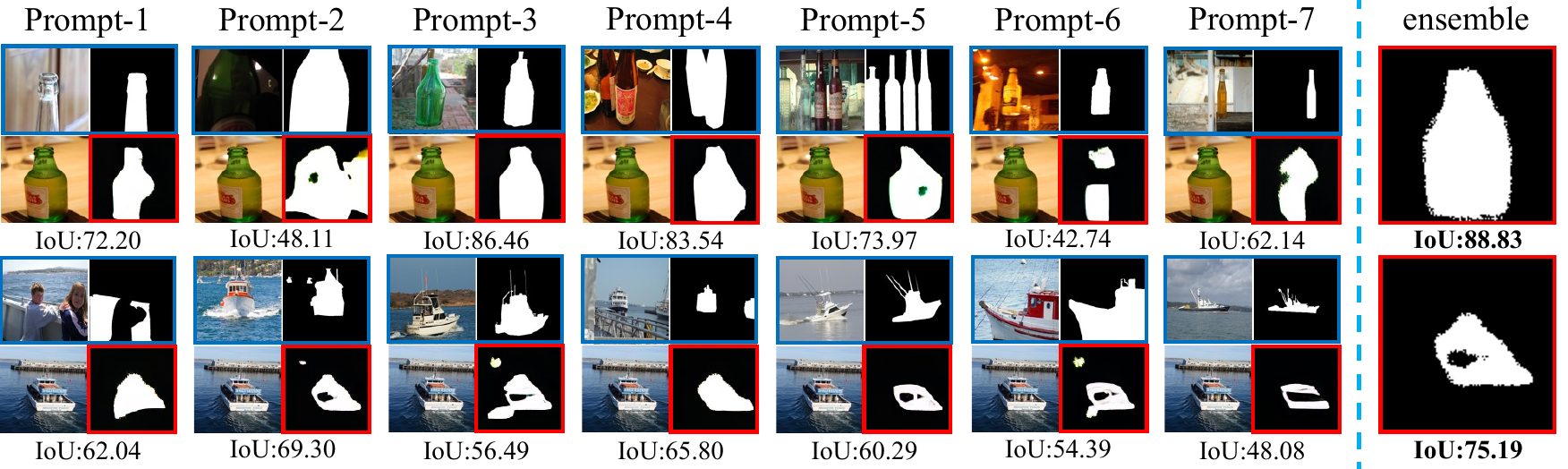}
	
	\caption{Visualization results of more prompts and ensemble. Increasing the number of prompts can significantly increase performance. Note that the \textcolor{red}{red box} in the figure is the prediction result, and the \textcolor{blue}{blue box} is the prompt pair. }
	\label{fig:more_prompt}
	\vspace{-0.4em}
\end{figure*}

 \begin{figure}
	\centering
	\includegraphics[width=\linewidth]{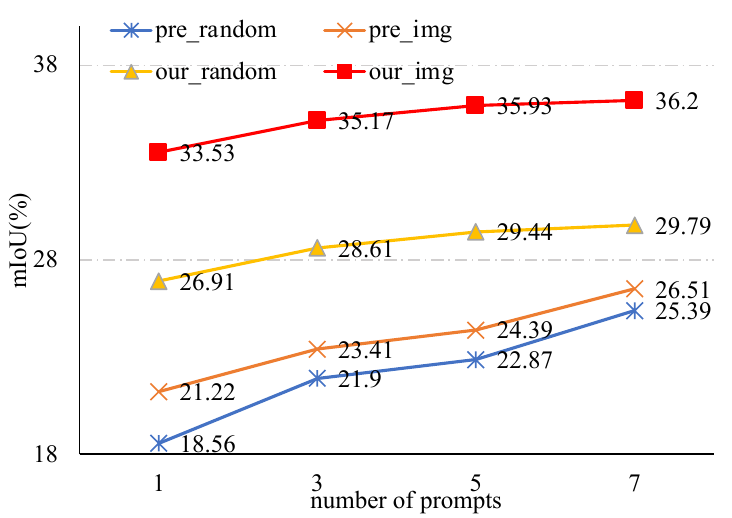}
	
	\caption{Ablation study of different methods on introduce more prompts. \textit{pre} means the method in VP~\cite{barvisual} and VPR~\cite{zhang2023makes}. \textit{random} and \textit{img} respectively represent the prompt selection method of random and image-level retrieval. }
	\label{fig:more_prompt_results}
	\vspace{-0.6cm}
\end{figure}

\vspace{-0.3em}
\noindent \textbf{Alleviating domain shift.} Domain shift is one of the key issues in visual in-context learning, as it can significantly impact the performance of the model. To alleviate the issue of domain shift, it is important to improve the robustness of the model to ensure its performance remains consistent and accurate in different environments. One approach to evaluating the robustness of a model is to test it on datasets that have different data distributions than the training dataset. We then investigate the robustness of prompt-SelF by switching the prompt dataset. Table~\ref{tab:domain} presents the experimental results, where \textit{"Pascal$\rightarrow$ Pascal"} denotes that both the query and prompt datasets are from Pascal, \textit{"COCO$\rightarrow$ Pascal"} denotes that the prompt dataset is from MSCOCO~\cite{lin2014microsoft} and the query dataset is from Pascal. The experiment results indicate that the pixel-level retrieval method can better mitigate the domain shift issue than the image-level retrieval methods. Moreover, in the \textit{"COCO$\rightarrow$ Pascal"} scenario, prompt-SelF achieved the best performance, indicating its robustness. The results indicate that prompt selection and prompt fusion are crucial elements that impact the robustness of visual context learning, and optimizing them can complementarily enhance the overall performance and robustness of the model.\\

\noindent \textbf{More prompts.} The previous method~\cite{barvisual,zhang2023makes} to increase the number of prompts involved creating a larger grid by dividing the $224\times 224$ space into 16 grids to obtain up to 7 prompt pairs. However, this approach significantly reduces the resolution of sub-images, which reduces the effectiveness of each prompt and negatively impacts visual in-context learning. In contrast, we propose a straightforward approach of combining the results from different prompts. Specifically, we sum the multiple prediction results and set a threshold of 0.5 to determine the final prediction result. 

Next, we conducted a comparison between the previous method and our proposed method on the impact of increasing the number of prompts on visual in-context learning. For this comparison, we selected two prompt selection methods: random and image-level retrieval, and the prompt fusion method \textbf{A1}. The results are presented in Figure~\ref{fig:more_prompt_results}, which shows that increasing the number of prompts positively affects visual in-context learning for all methods. However, previous methods significantly decrease the resolution of sub-images, resulting in 7 prompt pairs yielding worse results than our method with only 1 prompt. The visualization results in Figure~\ref{fig:more_prompt} demonstrate that different prompts can have a significant impact on the prediction results. When a bad prompt is introduced, it negatively affects the prediction result. However, with the ensemble strategy, the negative impact of the bad prompt can be alleviated, leading to more robust prediction results.

\section{Conclusion}
This paper investigates the key factors affecting visual in-context learning, namely prompt selection and prompt fusion. We demonstrate that high-quality prompts can significantly improve the performance of visual in-context learning, and prompt fusion is essential for utilizing the full knowledge in large-scale visual models. Based on the above insights, we propose a simple framework, prompt-SelF, which utilizes pixel-level retrieval for prompt selection and ensemble prediction results from multiple prompt fusions to fully exploit the knowledge in the large model. After conducting a large number of experiments, we have demonstrated the effectiveness of prompt-SelF in addressing the two factors of prompt selection and prompt fusion that affect visual in-context learning. Notably, the impact of prompt selection and prompt fusion on visual in-context learning is independent, and improving the ability of visual in-context learning requires optimizing both of these factors.

{\small
\bibliographystyle{ieee_fullname}
\bibliography{egbib}
}
\clearpage

\appendix
\vspace{-1.0em}
\noindent
{\Large {\textbf{Appendix}}}
\section{More Factors}
In this section, we undertake a comprehensive analysis of additional factors that impact the performance of visual in-context learning.

\subsection{The input-label mapping}
The input-label mapping for prompts plays a crucial role in NLP's in-context learning~\cite{brown2020language,ouyangtraining}. Prompts serve as the model's input, and the effectiveness of the model's understanding and processing of these inputs is largely determined by the quality of the input-label mapping. A well-designed input-label mapping can enable the model to encode the inputs accurately and produce precise outputs. Thus, in order to achieve optimal performance, it is necessary to meticulously design and optimize the input-to-label mapping for prompts. Similarly, we strongly contend that the input-label mapping of prompts plays a critical role in visual in-context learning. As a result, we conducted an experiment to substitute the correct label of the prompt with an incorrect one, utilizing the prompt selection methods of random and image-level retrieval, as well as the arrangement method of \textbf{A1}. Table~\ref{tab:mapping} presents the experimental results, which demonstrate the significant impact of input-label mapping on visual in-context learning. These findings indicate that an accurate input-label mapping can assist the model in comprehending new tasks with greater precision.

\begin{table}[]\footnotesize
\centering
\setlength\tabcolsep{1.5pt}
\renewcommand\arraystretch{1.3}
\caption{Ablation study of the input-label mapping. \protect\usym{2717} and \protect\usym{2713} indicate, respectively, whether the label corresponds to the prompt image. The best results based in-context learning are show in \textbf{bold}.}
\label{tab:mapping}
\begin{tabular}{c|c|cccc|c}
\hline
prompt selection                       & correct label & Fold-0 & Fold-1 & Fold-2 & Fold-3 & Means \\ \hline
\multirow{2}{*}{random}      &    \protect\usym{2713}      & 27.49 & 30.21 & 25.88  & 24.06 & \textbf{26.91} \\
                             &    \protect\usym{2717}      & 22.16 & 26.24 & 22.45  & 19.60 & 22.61\textcolor{red}{$_{(-4.30)}$}\\ \hline
\multirow{2}{*}{image-level} &     \protect\usym{2713}     & 34.70  & 35.92  & 32.92  & 31.12  & \textbf{33.53} \\
                             &     \protect\usym{2717}     & 27.81  & 31.37  & 26.64  & 25.76  & 27.90\textcolor{red}{$_{(-5.63)}$} \\ \hline
\end{tabular}
\vspace{-0.4cm}
\end{table}

\subsection{Feature Extractor in Prompt Selection}
The feature extractor used in prompt selection aims to extract features from both the prompt image and query image, and subsequently identify the most appropriate prompt for the query image based on these features. Therefore, it is imperative that the feature extractor can effectively convey semantic information across different regions of the image. In order to achieve this, we conducted experiments utilizing different feature extractors for both the pixel-level prompt selection and the prompt fusion method of A1. In order to conduct a comprehensive comparison, we selected three commonly-used feature extractors - VIT~\cite{dosovitskiyimage}, Beit~\cite{baobeit}, and CLIP~\cite{radford2021learning} - and the experimental results are presented in Table~\ref{tab:BB_PS}. The results indicate that the multi-modal feature extractor contains a greater amount of semantic information and can more effectively accomplish the task of prompt selection.

\begin{table}[]\footnotesize
\centering
\setlength\tabcolsep{3.5pt}
\renewcommand\arraystretch{1.3}
\caption{Ablation study of the feature extractor in Prompt Selection. The best results based in-context learning are show in \textbf{bold}.}
\label{tab:BB_PS}
\begin{tabular}{c|c|cccc|c}
\hline
Prompt Selection             & Weight & Fold-0 & Fold-1 & Fold-2 & Fold-3 & Means \\ \hline
\multirow{3}{*}{pixel-level} & VIT    & 35.66  & 37.71  & 33.31  & 33.50  & 35.05 \\
                             & Beit   & 34.55  & 37.23  & 32.66  & 32.83  & 34.32 \\
                             & CLIP   & \textbf{36.42}  & \textbf{38.47} & \textbf{34.56}  & \textbf{34.12}  & \textbf{35.89} \\ \hline
\end{tabular}
\vspace{-0.4cm}
\end{table}

 \begin{figure*}
	\centering
	\includegraphics[width=\linewidth]{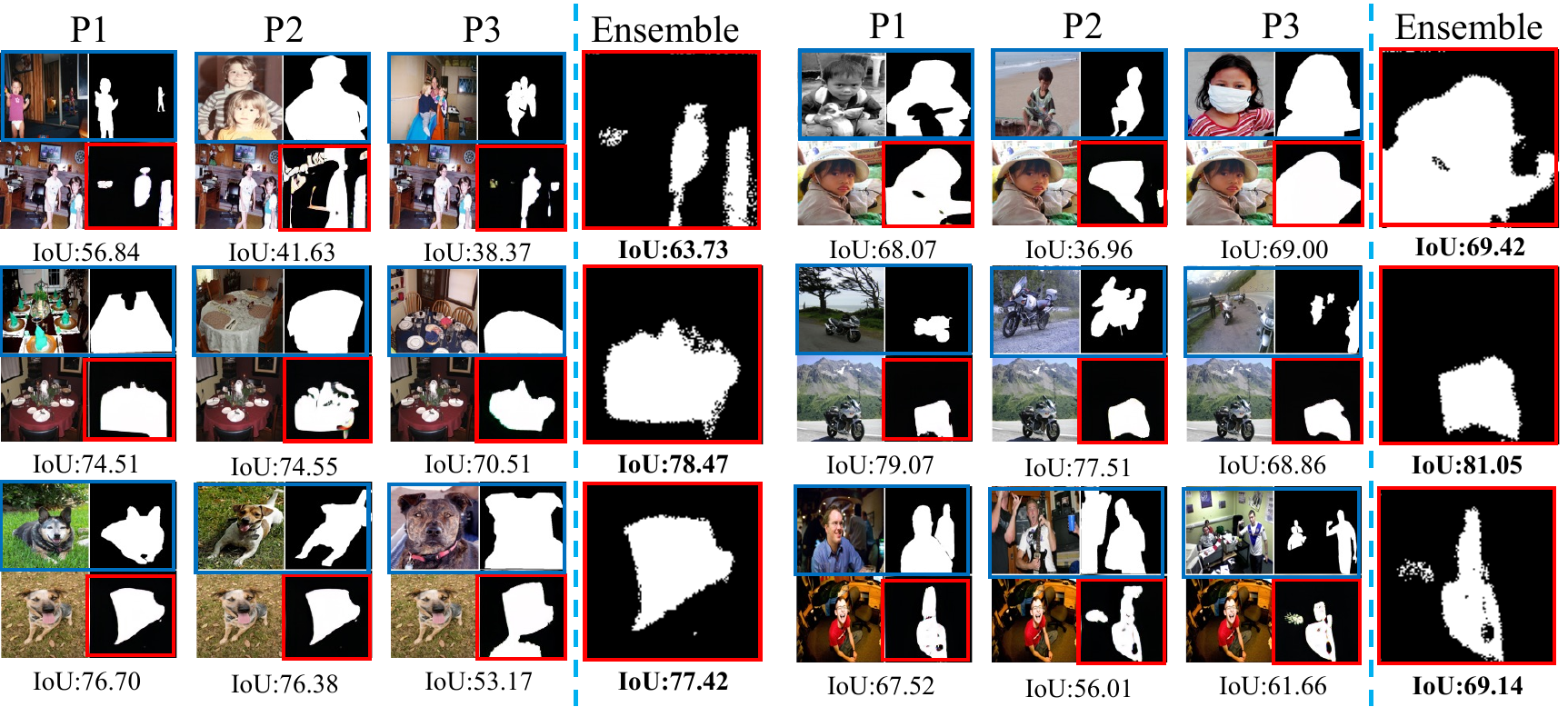}
	
	\caption{Visualization results of three prompts and ensemble. Increasing the number of prompts can significantly increase performance. Note that the \textcolor{red}{red box} in the figure is the prediction result, and the \textcolor{blue}{blue box} is the prompt pair.}
	\label{fig:apeendix_more3}
	\vspace{-0.4cm}
\end{figure*}
\vspace{-0.2cm}
\subsection{The Large-scale Visual Model}
Undoubtedly, enhancing the capacity of large-scale visual models can greatly enhance visual in-context learning. Nevertheless, the majority of current visual large-scale models~\cite{dosovitskiyimage,chen2022context,baobeit} lack the capacity for visual in-context learning. This is primarily due to the fact that these large-scale models are trained to comprehend the content of individual images, and therefore, are unable to directly capture the connections between different images. In light of this, prior research~\cite{barvisual,wang2022images} has endeavored to design large-scale visual models equipped with the capacity for visual in-context learning. However, given the nascent nature of visual in-context learning, current research is insufficient to facilitate experiments aimed at assessing the impact of different large-scale visual models on in-context learning. Consequently, we anticipate that future research will produce more robust and insightful work on the subject, thus enabling a more comprehensive understanding of the influence of various large-scale visual models on visual in-context learning.

\vspace{-0.3cm}
\section{More Visualization Results}
\begin{figure*}
	\centering
	\includegraphics[width=\linewidth]{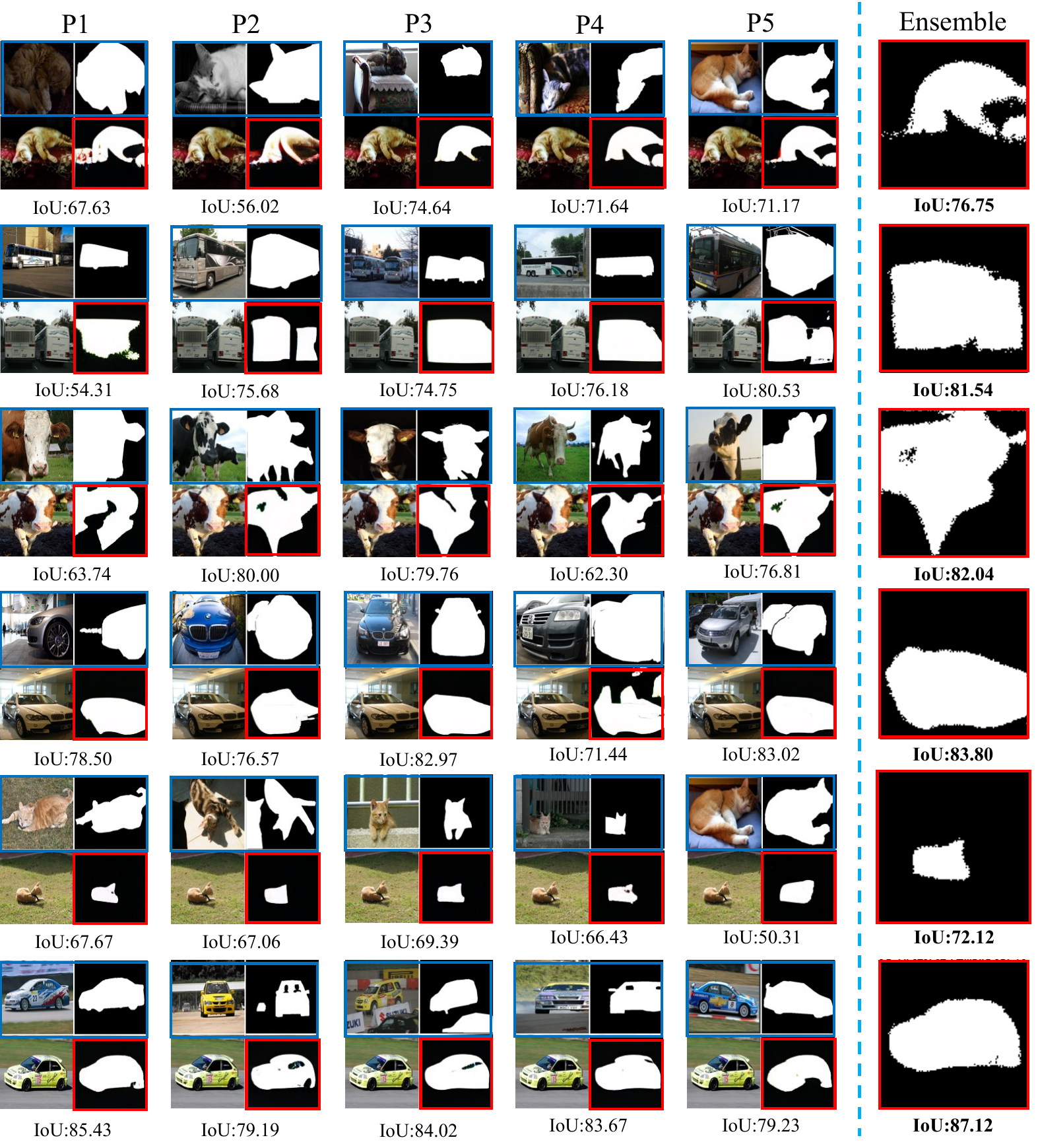}
	
	\caption{Visualization results of five prompts and ensemble. Increasing the number of prompts can significantly increase performance. Note that the \textcolor{red}{red box} in the figure is the prediction result, and the \textcolor{blue}{blue box} is the prompt pair.}
	\label{fig:apeendix_more5}
\end{figure*}

\begin{figure*}
	\centering
	\includegraphics[width=\linewidth]{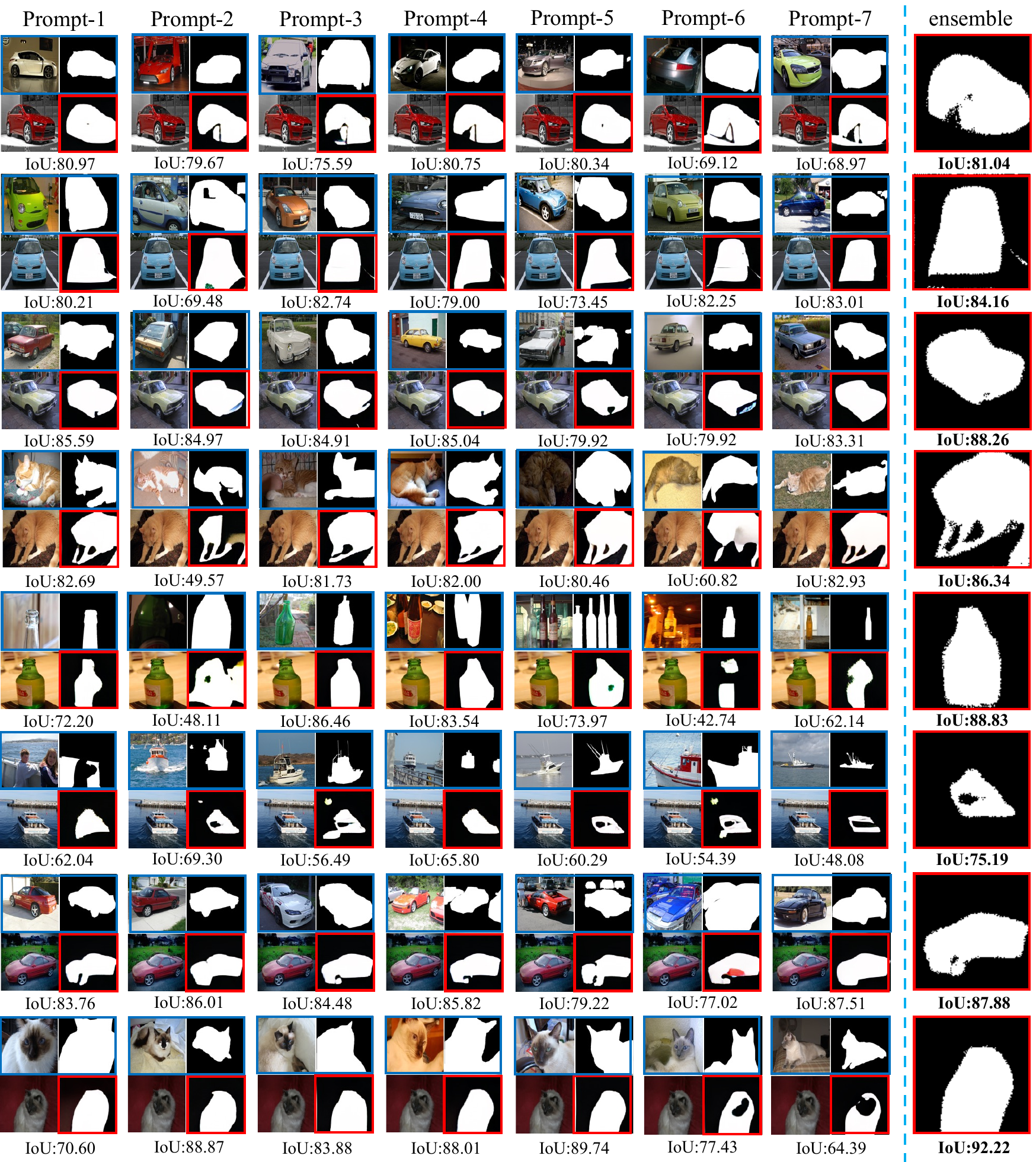}
	
	\caption{Visualization results of seven prompts and ensemble. Increasing the number of prompts can significantly increase performance. Note that the \textcolor{red}{red box} in the figure is the prediction result, and the \textcolor{blue}{blue box} is the prompt pair.}
	\label{fig:apeendix_more7}
\end{figure*}

\begin{figure*}
	\centering
	\includegraphics[width=\linewidth]{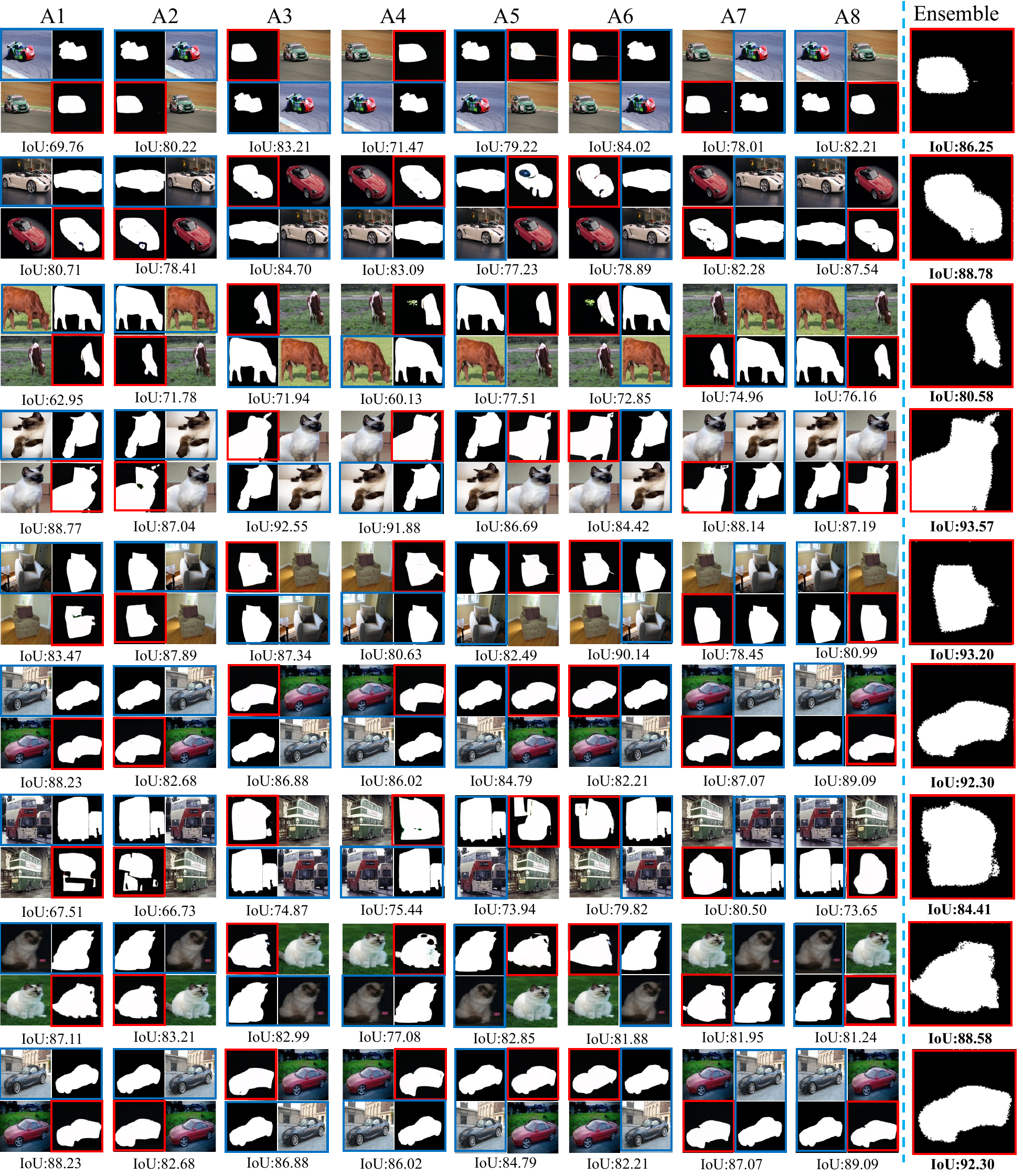}
	
	\caption{Visualization results of different prompt fusion methods. A1-A8 represent different prompt fusion methods, where the \textcolor{blue}{blue box} is the prompt pair, and the \textcolor{red}{red box} is the prediction result. The best results are show in \textbf{bold}.}
	\label{fig:appendix_diff_p}
\end{figure*}

To better visualize the benefits of different prompt fusion methods and more prompts for visual in-context learning, we present additional visualization results. Specifically, Figures~\ref{fig:apeendix_more3}, ~\ref{fig:apeendix_more5}, and ~\ref{fig:apeendix_more7} depict the impact of incorporating 3, 5, and 7 prompts, respectively, on the final outcome. Moreover, these visualizations enable us to observe the effect of different prompts on the prediction of a single query image. These results demonstrate the critical role of prompt selection in visual in-context learning and highlight how amalgamating multiple predictions can substantially enhance the accuracy of query image predictions. Furthermore, the impact of different prompt fusions is illustrated in Figure~\ref{fig:appendix_diff_p}. This further confirms that the arrangement of prompts and query images has a significant effect on the prediction results. Summarizing these results can greatly improve the performance of visual in-context learning.

\end{document}